\documentclass[11pt]{article}

\usepackage{array} 
\usepackage{tabularx}   
\usepackage{booktabs}   

\usepackage[preprint]{acl}

\usepackage{times}
\usepackage{latexsym}
\usepackage[most]{tcolorbox}
\usepackage[draft]{minted}
\usepackage[T1]{fontenc}
\usepackage{booktabs}
\usepackage{multirow}
\usepackage{xcolor}
\usepackage{amsmath}
\usepackage{amsfonts}

\usepackage[utf8]{inputenc}

\usepackage{microtype}

\usepackage{inconsolata}

\usepackage{graphicx}

\title{Seeing vs. Believing: Evaluating the Language Bias of Open-Source MLLMs in Counter-Intuitive Scenes}

\author{
  Chen Ling$^{1}$\thanks{These authors contributed equally to this work.},~
  Tongwei Zhang$^{2}$\footnotemark[1],~
  Hanqian Li$^{3}$\footnotemark[1],~
  Nai Ding$^{1}$\thanks{Corresponding author.}\\
  $^{1}$ College of Biomedical Engineering \& Instrument Science, Zhejiang University \\
  $^{2}$ Beijing University of Posts and Telecommunications \\
  $^{3}$ Hong Kong University of Science and Technology (Guangzhou)
}
\usepackage[capitalize]{cleveref} 
\begin{document}
\maketitle
\begin{abstract}
Multimodal Large Language Models (MLLMs) have demonstrated remarkable performance in mainstream visual understanding tasks, but their ability to process action scenes that contradict everyday common sense remains undertested. To address this gap, we introduce CAIT, a benchmark comprising 400 high-fidelity synthetic scenes focused on counter-intuitive visual actions, such as ``a rabbit is chasing a tiger'', where visual evidence explicitly contradicts common-sense expectations. We evaluate human, leading proprietary models (e.g., Claude and Gemini), and 14 representative open-source MLLMs. Humans achieve near-perfect performance (around 0.95 accuracy) and proprietary models demonstrate robust understanding (achieving up to 0.88 accuracy), standard open-source instruction-tuned models perform at the chance level. Further analysis demonstrates that this failure is driven by a strong language prior: rather than trusting the visual input, they automatically override the anomalous visual signals with statistically common text descriptions. Although introducing Chain-of-Thought reasoning mechanisms can improve accuracy, it significantly slows down the response and generates a new failure mode: models overthink the scenario and refuse to accept the actual visual content simply because it violates real-world physical laws. Finally, we demonstrate that targeted fine-tuning and structured prompting can effectively mitigate this reliance on language priors, enabling open-source models to accurately ground their reasoning in actual visual evidence. Our dataset is publicly available at \url{https://huggingface.co/datasets/LukeLing/CAIT}.
\end{abstract}
\begin{figure}[h]
    \centering
    \includegraphics[width=0.98\linewidth]{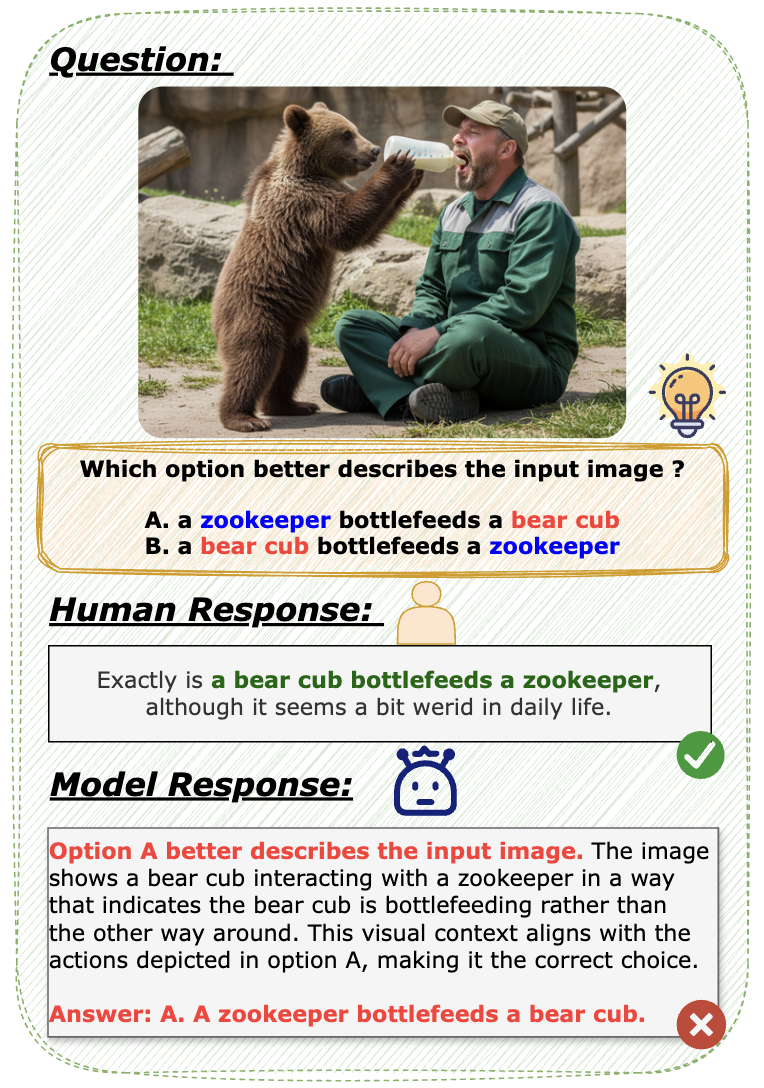}
    \caption{Example of an counter-intuitive action scene. Human responses comes from one of the annotators and Model Response is generated by GPT-4o-mini.}
    \label{fig:rabbit_drag_tiger}
\end{figure}

\section{Introduction}
Multimodal Large Language Models (MLLMs) have made remarkable breakthrough in multi-modal understanding tasks. They have shown excellent performance in tasks like Question Answering (QA)\citep{fan2026lfqae,min2026standardized}, Visual Question Answering (VQA) \citep{goyal2017vqav2,xun2026rtv, tao2025moss, shi2026androtmem, 11404921}, Image Captioning, and Visual Common Sense Reasoning \citep{zellers2019vcr}. Despite their impressive generalization abilities across large-scale standard datasets, their reasoning capabilities in counter-intuitive action scenes are undertested.



The disparity between human and model performance in these scenarios can be understood through \textbf{Dual Process Theory} \citep{kramer2014thinking}, which posits two modes of thought:
\begin{itemize}
    \item \textbf{System 1} (Intuition): A fast, automatic, and "blink-level" judgment mode.
    \item \textbf{System 2} (Deliberation): A slow, logical, and computationally expensive reasoning mode
\end{itemize}

For humans, identifying a counter-intuitive action—such as ``a bear cub bottlefeeds a zookeeper'' in Figure\ref{fig:rabbit_drag_tiger}—is primarily a System 1 task. Humans possess a "neural reflex" for interaction anomalies, allowing them to identify logical or physical violations almost instantaneously. This process relies on a deep, implicit accumulation of social and cultural common sense\citep{huang2025hyperg,hu2025videomark}.

In contrast, while current MLLMs are increasingly designed with analogous architectures \citep{Hagendorff_2023}, they fail to replicate human efficiency. Even with System 2 thinking, 
they still often rely on language patterns for superficial associations rather than truly understanding the semantic relationships or action logic between objects in image \citep{shah2022vlmbias}\citep{niu2021counterfactualvqacauseeffectlook}. Numerous studies point out the vulnerabilities of MLLMs in semantic reversal scenarios \citep{thrush2022winogroundprobingvisionlanguage} and common sense conflict scenarios.\citep{bittonguetta2023breakingcommonsensewhoops}
However, they mostly focus on object-level counterfactuals or text semantic conflicts, with little attention given to deep visual reasoning at the action level.


To address this, we introduce \textbf{CAIT} (\textbf{C}ounter-Intuitive \textbf{A}ction \textbf{I}mage-\textbf{T}ext dataset), a multimodal VQA benchmark specifically designed to evaluate whether MLLMs can overcome language priors to comprehend counter-intuitive action logic. Utilizing image synthesis pipeline, we construct scenarios that challenge models to move beyond pattern matching and perform genuine object-relationship reasoning. Our main contributions are as follows:

\begin{itemize}
    \item We present a novel VQA benchmark that systematically evaluates MLLMs' understanding of counter-intuitive visual actions and object-interaction relationships.
    \item We propose a semi-automated pipeline using state-of-the-art text-to-image model to generate high-fidelity, semantically consistent images with fine-grained VQA annotations.
    \item We provide an in-depth analysis of why current open-source MLLMs fail on CAIT and offer preliminary strategies
    to improve model performance in these challenging scenarios.
\end{itemize}
\begin{figure}[h]
    \centering
    \includegraphics[width=0.98\linewidth]{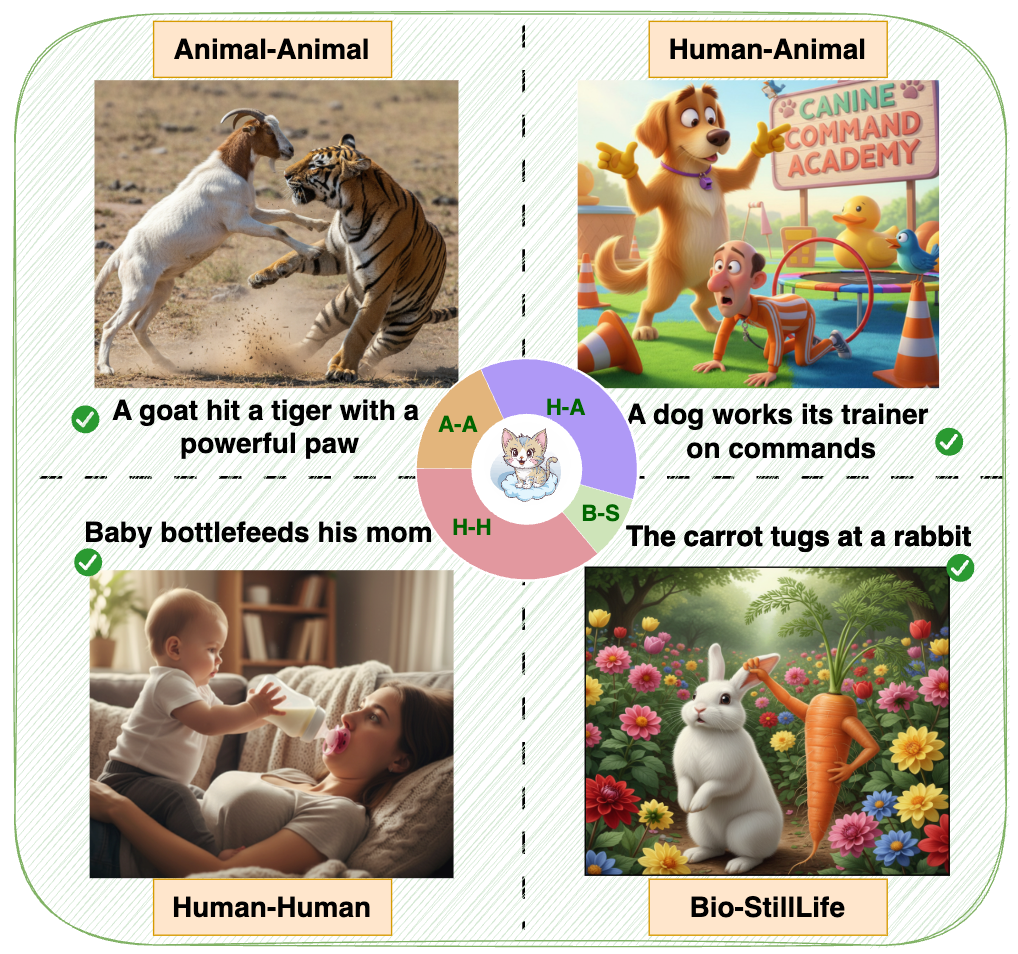}
    \caption{The four main classifications and corresponding examples of \textbf{CAIT}.}
    \label{fig:construction_framework}
\end{figure}

\begin{figure*}[h]
    \centering
    \includegraphics[width=0.90\linewidth]{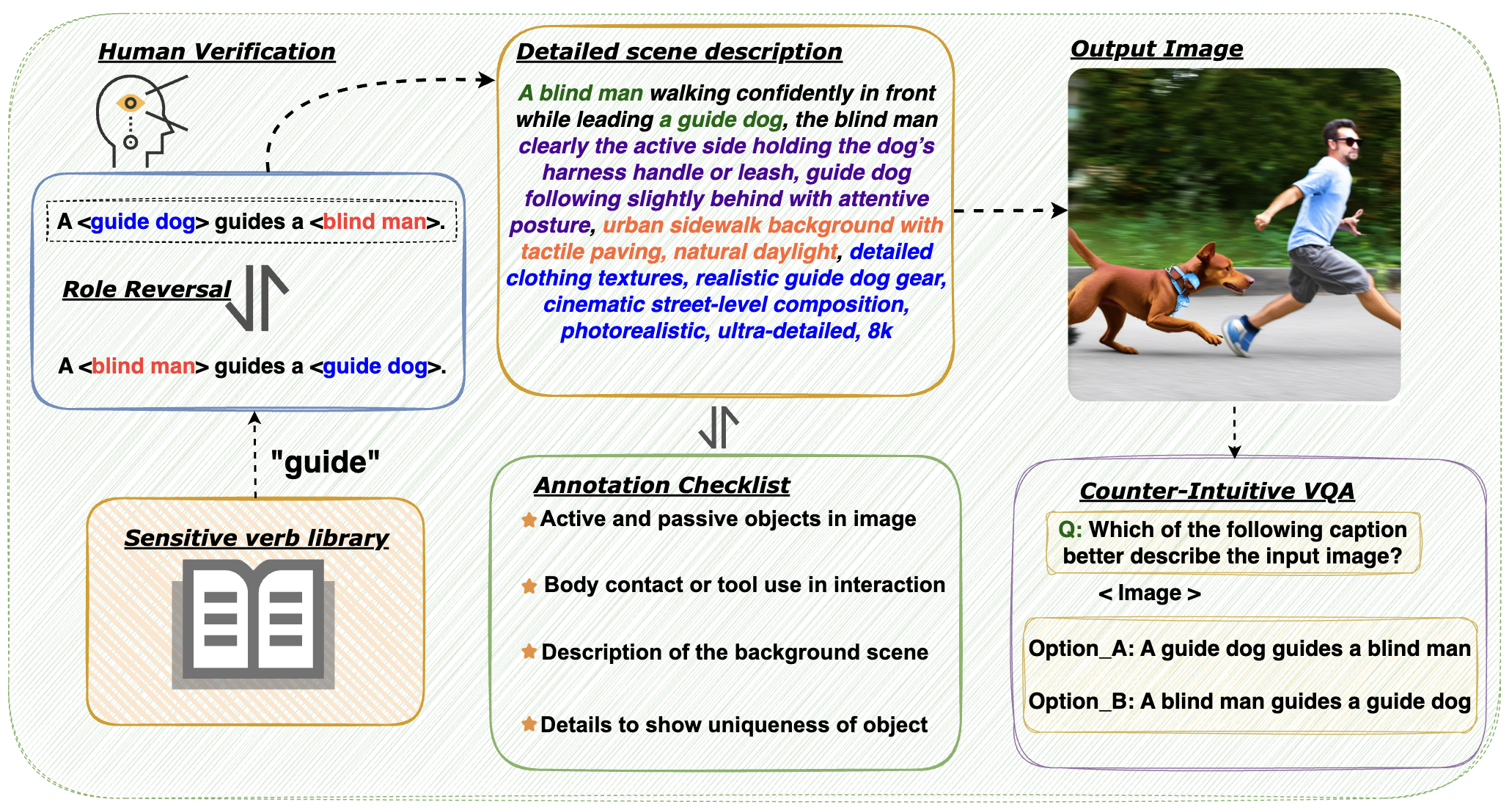}
    \caption{The construction framework of an counter-intuitive action image-text pair.}
    \label{fig:construction_framework}
\end{figure*}

\section{Related Work}


\paragraph{Evolution of MLLMs}
The field of multimodal reasoning has shifted significantly toward integrating Large Language Models (LLMs) with visual encoders to form MLLMs. Early breakthrough such as Flamingo \citep{alayrac2022flamingovisuallanguagemodel}, demonstrated remarkable few-shot \citep{brown2020languagemodelsfewshotlearners} capabilities in cross-modal reasoning through gated cross-attention mechanisms. Building on this, the current research landscape is dominated by models like LLaVA \citep{liu2023visualinstructiontuning}, Qwen-VL \citep{bai2023qwenvlversatilevisionlanguagemodel}, and InstructBLIP \citep{dai2023instructblipgeneralpurposevisionlanguagemodels}, which utilize visual feature projection and large-scale instruction tuning to align visual signals with the semantic space of LLMs. More recently, a new generation of "Thinking" MLLMs has emerged, such as 
Qwen-VL-Thinking series \citep{bai2025qwen3vltechnicalreport}. These models aim to simulate deliberative cognition by incorporating internal monologues or Chain-of-Thought (CoT) \citep{wei2023chainofthoughtpromptingelicitsreasoning} processes before generating final outputs, specifically targeting complex logical reasoning tasks.

\paragraph{Counterfactual Reasoning Benchmarks}
Counterfactual reasoning are essential for verifying whether MLLMs genuinely understand visual content or merely exploit statistical biases. To address the reliance on language priors, Winoground \citep{thrush2022winogroundprobingvisionlanguage} assesses visual-text alignment under semantic inversion by constructing matching tasks with semantically swapped text descriptions. Contrastive evaluation methods such as VQA-CP \cite{agrawal2018dontjustassumelook} construct uncommon distribution by disrupting the answer distribution consistency between training and test sets. However, these methods do not strictly involve genuine uncommon image scenes. More recently, COCO-Counterfactual \citep{le2023cococounterfactualsautomaticallyconstructedcounterfactual} tests model sensitivity to visual changes by modifying objects or attributes within images, while CFMM \citep{li2025lookdecidepromptingactive} evaluates reasoning capabilities through systematic counterfactual rewriting of question text, but they are predominantly limited to single-modality counterfactual construction.

\paragraph{Commonsense Conflict and Interaction Logic}
While benchmarks like WHOOPS! \citep{bittonguetta2023breakingcommonsensewhoops} introduce visual scenes that violate general commonsense, they often require complex, knowledge-intensive inference. Most related to our work is BLINK \citep{fu2024blinkmultimodallargelanguage}, which evaluates "blink-level" visual reasoning tasks. However, BLINK primarily targets core visual tasks such as spatial relations and counting. Our work fills a distinct gap by focusing specifically on interaction logic—the fundamental physical and social relationships between objects. Unlike WHOOPS!, which may rely on external cultural knowledge, CAIT emphasizes intuitive judgments derived from the agent-action-patient structure within a static frame. This focus on counter-intuitive actions allows for evaluation of how models balance visual signals against entrenched language priors.

\section{Methodology}


\begin{table*}[htbp]
\centering
\begin{tabular}{lcl}
\toprule
\textbf{Category}  & \textbf{Ratio} & \textbf{Key Characteristics} \\
\midrule
\multirow{2}{*}{Human-Human} &  22.00\% & Inversion of institutional power \\
 & 14.25\% & Reversal of caretaking duties across generations \\
\midrule
\multirow{2}{*}{Human-Animal}  & 22.25\% & Animals replacing humans as dominant figures \\
  & 14.00\% & Animals providing domestic care to humans \\
\midrule
Animal-Animal  & 18.00\% & Transposition of traditional predator and prey roles \\
\midrule
Bio-StillLife  & 9.50\% & Inanimate objects gaining autonomy \\
\bottomrule
\end{tabular}
\caption{Taxonomy and distribution of counter-intuitive scenes in the CAIT dataset.}
\label{tab:taxonomy_of_CAIT}
\end{table*}

\subsection{Overview}

To systematically evaluate MLLM reasoning, we carefully construct CAIT through a generation pipeline that pairs high-fidelity images with rigorous semantic annotations.

\paragraph{Diverse Visual Action Scenes}

CAIT consists of 400 distinct scenarios organized into four taxonomic categories and six characteristic variations. Distribution of the scenes is detailed in Table \ref{tab:taxonomy_of_CAIT}.

\paragraph{Instant Thinking Mode}
The task is formulated to evaluate "neural reflex" perception, prioritizing immediate visual recognition over multi-step causal analysis. Unlike standard VQA, CAIT requires models to determine commonsense alignment directly from visual cues, bypassing the linguistic parsing of hypothetical conditions.

\paragraph{Blink-Level Tasks for Human}
As illustrated in Figure \ref{fig:rabbit_drag_tiger}, the images in CAIT are designed to achieve "blink-level" difficulty. This establishes a human baseline where interaction anomalies are immediately perceptible, typically requiring few seconds for accurate identification.

\paragraph{High-Quality Synthetic Data}

We leverage Gemini-2.5-flash-image \citep{comanici2025gemini25pushingfrontier} to generate images, which minimizes data contamination risks associated with web-crawled datasets. Each sample undergoes human verification to maintain high semantic fidelity and visual clarity.

\subsection{Data Processing}
The data processing pipeline is divided into three phases: Verb Library Filtering, Counter-intuitive Image-Text Generation, and VQA Dataset Design.

\paragraph{Verb Library Filtering}

Using VerbNet \citep{kipper-schuler-etal-2009-verbnet} as a foundation, we filter verbs based on four core principles, see detailed examples in Appendix \ref{apx:verb_library_filtering}:
\textbf{1)} \textbf{Semantic Consistency:} Verbs must explicitly describe the action relationship between objects. 
\textbf{2)} \textbf{Semantic Non-interchangeability:} The verb must follow "Agent-Action-Patient" structure, 
    meaning that swapping the subject and object results in a fundamental change in the scenario's meaning or logic.
\textbf{3)} \textbf{Static Recognizability:} 
    Guided by existing visual action definitions\citep{ronchi2015describingcommonhumanvisual}, actions must be clearly discernible in a single static frame without temporal or contextual cues.
\textbf{4)} \textbf{Commonality:} Verbs should be high-frequency and general-purpose, strictly avoiding domain-specific jargon.

Following these principles, we construct the \textit{\textbf{Sensitive-Action Verb Library}}, consisting of 53 categories and a total of 318 verbs. Table \ref{tab:verb_library_classification} presents the verb classification and corresponding judgment.


\begin{table}[ht]
\begin{tabular}{lcc}
\hline
Class & Verb & Judgement Basis \\
\hline
6 & 40 & direction of agent \\
19 & 106 & active, passive, body contac \\
12 & 80 & active, passive, tool using \\
12 & 80 & active, passive, body part \\
4 & 21 & position of agent \\
\hline
\end{tabular}
\caption{Classification of \textit{Sensitive-Action Verb Library}}
\label{tab:verb_library_classification}
\end{table}

\paragraph{Counter-intuitive Image-Text Generation}
We begin by selecting verbs from the sensitive-action library to construct counter-intuitive texts. Specifically, standard sentences with a Agent-Verb-Patient structure typically convey intuitive and commonsense meanings commonly found in daily language. However, the selected verbs are semantically non-interchangeable. That is, swapping the 
agent and patient significantly alters the sentence's meaning, resulting in a counter-intuitive scenario.

To efficiently collect a larger scale of such sentences and their corresponding images, we designed a data synthesis framework 
as illustrated in Figure \ref{fig:construction_framework}.
    \textbf{1)} \textbf{Sentence Generation}: 
    We prompt GPT-4o-mini to generate semantic scene pairs derived from specific verbs (see Table~\ref{tab:prompt_step1}). For each verb, the model produces both a normal, commonsense sentence and a corresponding counter-intuitive version by swapping the subject and object, creating a logical but extremely uncommon scenario.
    \textbf{2)} \textbf{Human Verification}: 
    We implement a human-in-the-loop verification process using a custom GUI (see Table~\ref{tab:prompt_step2}). Human annotators review the raw generated pairs to filter out nonsensical inversions and configure essential parameters for each scene, specifically determining whether to anthropomorphize the subject and selecting the visual style (realistic or cartoon).
    \textbf{3)} \textbf{Scene Description}: 
    We then prompt GPT-4o-mini to generate highly detailed and visually grounded descriptions (see Table~\ref{tab:prompt_step3}). These descriptions are tailored to clearly convey the subject--object role reversal while adhering to the specified style and anthropomorphism settings.
    \textbf{4)} \textbf{Image Generation and Filtering}: 
    Finally, we employ Gemini-2.5-flash-image to generate images based on the refined scene descriptions (see Table~\ref{tab:prompt_step4}). Human annotators then rigorously review and filter the output to ensure each image-text pair clearly conveys the counter-intuitive concept while removing ambiguous or distracting elements.

\paragraph{VQA Dataset Design}
The resulting CAIT benchmark comprises 400 images paired with specific questions for MLLM evaluation. As illustrated in Figure\ref{fig:construction_framework}, each image is paired with a single question where the question stem remains constant, and only the answer choices vary. The two primary options consist of the original commonsense text and the counter-intuitive text where the subject and object are swapped.

\subsection{Human Annotation for VQA}
\label{subsec:human_annotation}
We invited five annotators to perform manual annotations, establishing a high-quality comparison baseline for existing MLLMs. 
The annotation process records the time interval between the initial presentation of a question and the final submission as the reasoning time. Experimental results demonstrate that even if participant over 50 years old can achieve an average reasoning time of around 5 seconds. This consistently validates that the CAIT dataset achieves its "blink level" design goal, allowing humans to make near-instantaneous judgments. 
Detailed information is provided in Appendix \ref{apx:annotator}.

\begin{table*}[t]
\centering
\begin{tabular}{lcccccc}
\toprule
\multirow{2}{*}{\textbf{Model}} & \multirow{2}{*}{\textbf{Param}} 
& \multicolumn{3}{c}{\textbf{Counter-intuitive VQA}} 
& \multicolumn{2}{c}{\textbf{LLM-re-Eval}} \\
\cmidrule(lr){3-5} \cmidrule(lr){6-7}
 &  & Acc & F1 & Time & Acc & F1 \\
\midrule

\multicolumn{7}{l}{\textbf{\textit{Upper Bound}}} \\
Human Annotators & -- & 0.948 & 0.948 & 3.740 & -- & -- \\
GT Prompt & -- & -- & -- & -- & 0.984 & 0.985 \\

\midrule
\multicolumn{7}{l}{\textbf{\textit{Leading proprietary MLLMs}}} \\
Gemini-2.5-pro & -- & 0.885 & 0.886 & 0.804 & 0.835 & 0.835 \\
Claude-4.6-opus & -- & 0.863 & 0.863 & 0.817 & 0.788 & 0.788 \\
GPT-5.4 & -- & 0.805 & 0.804 & 0.832 & 0.778 & 0.778 \\
Grok-4.3 & -- & 0.893 & 0.892 & 0.819 & 0.788 & 0.789 \\

\midrule
\multicolumn{7}{l}{\textbf{\textit{Open-source Instruct MLLMs}}} \\
Qwen2.5-VL-Instruct & 7B & 0.420 & 0.347 & 0.940 & 0.660 & 0.660 \\
Qwen3-VL-Instruct & 8B & 0.380 & 0.327 & 0.853 & \textbf{0.762} & \textbf{0.762} \\
Gemma-3-it & 12B & 0.360 & 0.353 & 0.843 & 0.596 & 0.596 \\
LLaMA3.2-Vision & 11B & 0.375 & \textbf{0.372} & 2.622 & 0.607 & 0.607 \\
LLaVA-NeXT & 7B & \textbf{0.475} & 0.349 & 1.870 & 0.548 & 0.548 \\

\addlinespace
Qwen2.5-VL-Instruct & 3B & 0.373 & 0.300 & 0.924 & 0.584 & 0.585 \\
Qwen3-VL-Instruct & 4B & 0.328 & 0.326 & 0.633 & 0.707 & 0.707 \\
Gemma-3-it & 4B & 0.328 & 0.334 & \textbf{0.819} & 0.541 & 0.541 \\

\midrule
\multicolumn{7}{l}{\textbf{\textit{Open-source Thinking MLLMs}}} \\
Qwen3-VL-Thinking & 8B & 0.640 & 0.635 & 10.645 & 0.696 & 0.696 \\
GLM-4.1V-Thinking & 9B & 0.630 & 0.647 & \textbf{5.770} & \textbf{0.733} & \textbf{0.733} \\
Kimi-VL-A3B-Thinking & 16B & \textbf{0.683} & \textbf{0.680} & 7.396 & 0.689 & 0.689 \\
InternVL-3.5 & 8B & 0.545 & 0.545 & 106.512 & 0.655 & 0.655 \\

\addlinespace
Qwen3-VL-Thinking & 4B & 0.625 & 0.622 & 14.385 & 0.695 & 0.695 \\
InternVL-3.5 & 4B & 0.618 & 0.618 & 50.333 & 0.683 & 0.683 \\

\bottomrule
\end{tabular}
\caption{Performance of MLLMs on CAIT. The LLM-re-Eval scores are averaged over the three proprietary judges (Gemini-2.5-pro, Claude-4.6-Opus and GPT-5.4); the per-judge results are reported in Table \ref{tab:llm_re_eval} of Appendix \ref{apx:llm_re_eval_judges}. Values denoted in \textbf{bold} indicate the highest performance in each open-source model group.}
\label{tab:main_results}
\end{table*}

\section{Experiment}

\subsection{Model Selection}
To comprehensively evaluate the reasoning capabilities of MLLMs in CAIT scenarios, we selected leading proprietary models (Claude-4.6-opus and Gemini-2.5-pro) and 14 representative open-source models, categorized by their parameter scale and reasoning paradigms. 
\paragraph{Instruct MLLMs}
Instruct-based models are fine-tuned to follow human instructions directly, 
representing the System 1 thinking mode. Our selection includes 3B, 7B and 72B version of Qwen2.5-VL-Instruct \citep{bai2025qwen25vltechnicalreport}, 4B, 8B and 235B version of Qwen3-VL-Instruct \citep{bai2025qwen3vltechnicalreport}, 4B and 12B version of Gemma-3-it \citep{gemmateam2025gemma3technicalreport}, LLaVA-NeXT-7B \citep{li2024llavanextinterleavetacklingmultiimagevideo}, and Llama-3.2-Vision-11B\citep{grattafiori2024llama3herdmodels}.

\paragraph{Thinking MLLMs}
Thinking models are designed to simulate deliberative cognition by incorporating an explicit internal reasoning process before generating the final output, representing the System 2 thinking mode. We evaluate 4B and 8B version of Qwen3-VL-Thinking and InternVL-3.5, GLM-4.1V-9B-Thinking \citep{vteam2026glm45vglm41vthinkingversatilemultimodal} and Kimi-VL-A3B-Thinking\citep{kimiteam2025kimivltechnicalreport}.

\subsection{Implementation Details}
\label{Implementation Details}
We set up two tasks in order to evaluate not only the primary VQA performance, but also the semantic understanding consistency of MLLMs. To establish an ideal performance upper bound for the evaluation, we set reference baselines for both categories. In the Counter-intuitive VQA task, we utilize the average performance of human annotators (see \ref{subsec:human_annotation}) as the human baseline to measure the gap in models' intuitive judgment. For the Gemini-re-eval task, we use the \textbf{GT Prompt} as input and the judgment results of Gemini-2.5-pro as the ground-truth baseline to verify the reasoning upper bound at the semantic description level.


\paragraph{Counter-intuitive VQA} This task evaluates the model's capacity when confronted with visual actions that violate common sense. To provide a comprehensive assessment, we utilize Accuracy ($\text{Acc}$) and $F1$-score as the primary evaluation metrics:
\begin{equation}
\text{Acc} = \frac{1}{N} \sum_{i=1}^{N} \mathbb{I}(\hat{y}_i = y_i)
\end{equation}
\begin{equation}
F1 = \frac{2 \cdot \text{Precision} \cdot \text{Recall}}{\text{Precision} + \text{Recall}}
\end{equation}
where $N$ denotes the total number of test samples, $\hat{y}_i$ and $y_i$ represent the predicted and ground-truth options for the $i$-th instance. 

To deeply analyze computational overhead and logical depth, we precisely record the inference time ($t_{infer}$) for each question. To eliminate measurement errors caused by hardware variations, all open-source models are deployed on a single NVIDIA A100 (80GB) GPU using the vLLM framework \citep{kwon2023efficientmemorymanagementlarge} for online inference services. In the inference settings, we uniformly adopt a greedy decoding strategy with a temperature of \textbf{$T=0.0$} to ensure output consistency. This process can be directly expressed as:
\begin{equation}
\mathcal{A}_{direct} = f_{\text{MLLM}}(I, Q)
\end{equation}
where $I$ is the original image, $Q$ is the textual question, and $\mathcal{A}_{direct}$ is the directly output A/B option.

For Instruct MLLMs, in addition to the standard "instant answer", we introduce a CoT variant experiment to simulate the System 2 reasoning mode. In this setting, we guide the model to generate a reasoning chain before providing the final option. This aims to explore whether the failure of instruction-tuned models in these scenarios stems from initial misdirection at the visual perception layer or a lack of subsequent logical integration capabilities.

\paragraph{LLM-re-eval} To further deconstruct whether the model has truly captured the interaction semantics between different objects, we design an indirect reasoning experiment. The main purpose is to decouple image perception from logical judgment. First, the model under test generates a text description based on the image, which is then input alongside the question into one of three proprietary judges Gemini-2.5-pro, Claude-4.6-Opus, or GPT-5.4 for secondary evaluation:
\begin{equation}
\mathcal{D} = f_{\text{MLLM}}(I), \quad \mathcal{A}_{eval} = g_{\text{Judge}}(\mathcal{D}, Q)
\end{equation}
where $g_{\text{Judge}}$ represents the discriminative model and $\mathcal{D}$ is the generated semantic description, we took the average accuracy and F1 of three Judges as the LLM-re-eval.

Since all counter-intuitive images in the CAIT are synthesized by advanced text-to-image model, we define the original text prompt used to synthesize the image as the \textit{GT Prompt}. As the direct source of image generation, \textit{GT Prompt} represents the most accurate and unambiguous textual description of the action logic within that image.

Experimental results indicate that when any of the three proprietary judges take the \textit{GT Prompt} as input, average accuracy reaches approximately \textit{98\%}. Among the evaluated models, proprietary judges achieve the highest LLM-re-eval scores, while open-source Instruct and Thinking models exhibit substantially lower accuracies, highlighting challenges in capturing the semantic logic of counter-intuitive interactions. This result demonstrates three proprietary judges' superior semantic logic capabilities and verifies that the counter-intuitive VQA problem is logically resolvable provided the image description is sufficiently accurate. Consequently, a low score in LLM-re-eval directly reflects a model's failure to correctly identify counter-intuitive interaction relationships during the visual perception phase.
\begin{table}[htbp]
\centering
\resizebox{0.5\textwidth}{!}{
\begin{tabular}{lccc}
\toprule
Model & Accuracy & F1 Score & Infer Time \\
\midrule
\multicolumn{4}{l}{\textit{$\sim$10B Instruct MLLMs}} \\
Qwen2.5-VL-7B-Instruct & 0.483 & 0.472 & 3.060 \\
Qwen3-VL-8B-Instruct & 0.643 & 0.641 & 7.747 \\
Gemma-3-12b-it & 0.535 & 0.533 & 3.824 \\
LLama3.2-vision-11B & 0.498 & 0.509 & 7.713 \\
LLaVA-NeXT-7B & 0.458 & 0.452 & 3.537 \\
\midrule
\multicolumn{4}{l}{\textit{$\sim$3B Instruct MLLMs}} \\
Qwen2.5-VL-3B-Instruct & 0.485 & 0.468 & 2.296 \\
Qwen3-VL-4B-Instruct & 0.603 & 0.602 & 22.809 \\
Gemma3-4B-it & 0.478 & 0.477 & 1.666 \\
\bottomrule
\end{tabular}
}
\caption{Results of Instruct models with CoT reasoning.}
\label{tab:instruct_cot}
\end{table}

\subsection{Main Result}
Experimental results in Table \ref{tab:main_results} demonstrate that the performance of all evaluated open-source MLLMs is significantly lower than the upper bound when processing counter-intuitive action scenes. While proprietary models, show strong performance with accuracies up to 0.88 in VQA task, a substantial performance gap persists between open-source and proprietary models. This indicates that current open-source models still face substantial challenges in parsing CAIT scenarios.

The LLM-re-Eval scores in Table \ref{tab:main_results} are averaged over the three proprietary judges, whose per-judge breakdown is reported in Table~\ref{tab:llm_re_eval} (Appendix~\ref{apx:llm_re_eval_judges}). The GT-Prompt upper bound reaches around 0.98 across all judges. Proprietary models achieve the highest accuracies, while open-source Instruct models range from 0.532 to 0.773 and Thinking models from 0.663 to 0.743. Results are consistent across judges, reflecting judge-independent evaluation.

\paragraph{Language bias of Instruct MLLMs}
Instruct MLLMs exhibit rapid inference speeds, 
 reflecting System 1 fast-thinking traits, but the accuracy consistently perform at the chance level. This reveals that such models rely heavily on language priors. When confronted with conflicts between visual facts and common sense, they tend to overlook actual visual features in favor of common-sense biases, revealing a lack of genuine understanding regarding physical logic.

\paragraph{Gains and Costs of Thinking}
Experimental data clearly show that incorporating a thinking guidance mechanism effectively corrects erroneous judgments. For Instruct models in Table \ref{tab:instruct_cot}, performance generally improves upon the introduction of CoT reasoning. Native Thinking models similarly exhibit enhanced robustness. However, the average inference time per question is several times—or even orders of magnitude—higher than that of Instruct models. This suggests that such logical correction comes at a significant computational cost, highlighting the high inference premium of "slow thinking" in counter-intuitive reasoning.

\paragraph{Semantic Collapse in Gemini-re-Eval}
As demonstrated in \ref{Implementation Details}, the logic of the problem itself is not inherently difficult for a powerful MLLM. However, when the input is replaced with captions generated by open-source models (except Qwen3-VL series), the performance of Gemini-2.5-pro drops significantly. This suggests that when describing images, models often automatically "correct" the observed counter-intuitive facts into erroneous common-sense descriptions during output, leading to a collapse of visual signals during semantic descriptions transformation.
\section{Analysis}
\subsection{Performace across different domains}
We further analyze the impact of various subjects on model accuracy, with average results summarized in Table \ref{tab:avg_acc_across_domains} and details provided in Appendix \ref{apx:acc_across_domains}. The models generally outperform other domains within the Animal-Animal category, which may be attributed to the higher discriminability of visual features in predatory relationships at the lower levels of visual processing. Conversely, the Human-Human domain renders Instruct models particularly susceptible to being misled by entrenched social common-sense priors. 
Notably, a comparison between System 1 and System 2 models reveals that the Bio-StillLife domain yields the most substantial reasoning gain, demonstrating that explicit logical deduction in reasoning can effectively recover authentic visual signals from perceptual hallucinations in physically absurd scenarios.
\begin{table}[ht]
\centering
\begin{tabular}{lcccc}
\toprule
Model & H-H & H-A & A-A & B-S \\
\midrule
\multicolumn{5}{l}{\textit{Instruct MLLMs}} \\
$\sim$10B & 0.323 & 0.433 & 0.447 & 0.389 \\
$\sim$3B & 0.299 & 0.366 & 0.509 & 0.272 \\
\midrule
\multicolumn{5}{l}{\textit{Thinking MLLMs}} \\
$\sim$10B & 0.633 & 0.607 & 0.646 & 0.533 \\
$\sim$3B & 0.631 & 0.559 & 0.590 & 0.619 \\
\bottomrule
\end{tabular}
\caption{The average accuracy of open-source models across categories (A-A: Animal-Animal, B-S: Bio-StillLife,
H-A: Human-Animal, H-H: Human-Human).}
\label{tab:avg_acc_across_domains}
\end{table}


\subsection{Attempts to boost performance}
\paragraph{Structured Prompt for System 2 Enhancement}
\label{subsub:structured_prompt}
To elevate the reasoning ceiling of models in System 2 mode, we designed a Structured Prompt specifically tailored for scenarios in CAIT (see Appendix \ref{apx:structured_prompt}). This strategy is intended to guide models in explicitly analyzing the interaction relationships between objects, arriving at conclusions by questioning physical plausibility and deliberately evaluating option logic. Table \ref{tab:avg_result_on_structure_prompt} presents the average performance metrics, with results for individual models provided in Appendix \ref{apx:acc_structured_prompts_on_different_models}. The results indicate that nearly all models, with the exception of LLama3.2-vision-11B and Qwen2.5-VL-3B-Instruct, benefited from this strategy, with 10B-scale models demonstrating significant gains. Notably, Qwen3-vl series showed exceptional performance under structured prompting, with Qwen3-vl-thinking achieving a leap of nearly 14\%. This suggests that reinforcing deliberative instructional guidance can effectively unlock the latent reasoning potential of MLLMs.

\begin{table}[htbp]
\centering
\small 
\renewcommand{\arraystretch}{1.2} 
\begin{tabular}{lcc}
\toprule
\textbf{Model} & \textbf{Accuary} & \textbf{F1 Score} \\ 
\midrule
\multicolumn{3}{l}{\textit{Instruct Models}} \\
\quad $\sim$10B & 0.554 {\color{red} $\uparrow$ 4.2\%} & 0.546 {\color{red} $\uparrow$ 3.7\%} \\
\quad $\sim$3B  & 0.518 {\color{red} $\uparrow$ 1.3\%} & 0.518 {\color{red} $\uparrow$ 3.0\%} \\
\addlinespace 
\midrule
\multicolumn{3}{l}{\textit{Thinking Models}} \\
\quad $\sim$10B & 0.688 {\color{red} $\uparrow$ 7.2\%} & 0.687 {\color{red} $\uparrow$ 6.8\%} \\
\quad $\sim$3B  & 0.650 {\color{red} $\uparrow$ 3.8\%} & 0.648 {\color{red} $\uparrow$ 3.6\%} \\
\bottomrule
\end{tabular}
\caption{The average accuary and F1 score of open-source models after structured prompt.}
\label{tab:avg_result_on_structure_prompt}
\end{table}

\paragraph{LoRA-SFT for System 1 Calibration}
To address the vulnerability of instruction-tuned models to language-prior interference in System 1 mode, we employed Supervised Fine-Tuning to directly calibrate the models' cognitive preferences. We partitioned the CAIT into a 60\% training set and a 40\% test set, utilizing LoRA \citep{hu2021loralowrankadaptationlarge} to fine-tune 
Qwen2.5-vl-Instruct model while freezing the vision encoder. This ensures that the models learn new semantic correspondences while maintaining robust visual feature extraction capabilities. Experimental results in Table\ref{tab:lora_results} show that both fine-tuned models achieved performance increments of over 15\%. This proves that fine-tuning on domain-specific distributions can effectively mitigate reliance on common-sense language priors, enabling more accurate, near-instantaneous responses to counter-intuitive visual signals.


\begin{table}[htbp]
\centering
\small 
\renewcommand{\arraystretch}{1.3} 
\begin{tabular}{lcc}
\toprule
\textbf{Model} & \textbf{Accuracy} & \textbf{F1 score} \\ 
\midrule
Qwen-2.5-vl-7B & 0.425 & 0.322 \\
Qwen-2.5-vl-7B-LoRA & 0.575 {\color{red} $\uparrow$ 15.0\%} & 0.473 {\color{red} $\uparrow$ 15.2\%} \\
\addlinespace[0.5em] 
Qwen-2.5-vl-3B & 0.400 & 0.308 \\
Qwen-2.5-vl-3B-LoRA & 0.581 {\color{red} $\uparrow$ 18.1\%} & 0.498 {\color{red} $\uparrow$ 18.9\%} \\
\bottomrule
\end{tabular}
\caption{Comparison of Qwen2.5-vl series during SFT.}
\label{tab:lora_results}
\end{table}

\subsection{Failure Analysis}
We further investigate the failure cases from \ref{subsub:structured_prompt} and categorize them into four modes based on the source of hallucination:
\textbf{1)} \textbf{Statistical Frequency Bias:} The model acknowledges the possibility of an event but rejects valid visual evidence because the scenario contradicts the statistical likelihoods observed in its training distribution. \textbf{2)} \textbf{Logical Impossibility:} Model employs a rigid rejection mechanism, asserting that visual content violates fundamental physical laws or logical constraints, often explicitly labeling the scenario as ``absurd'' or ``impossible''. \textbf{3)} \textbf{Semantic Binding Failure:} A pure vision-language alignment error where the model correctly identifies objects but misattributes semantic roles (e.g., Agent-Patient reversal) or spatial relation due to attention failures. \textbf{4)} \textbf{Alignment Failure:} A disconnect between reasoning and generation, where the model deduces the correct answer in its chain-of-thought but hallucinates an inverted final label due to the re-emergence of language priors.
    
    
    
Figure \ref{fig:error_dist} shows the probability of each error. Also, we list the specific failure case in Appendix \ref{sec:error_analysis}. Instruct models are predominantly plagued by shallow \textit{Statistical Frequency Bias} and \textit{Spatial Binding Failures}, reflecting System 1's reliance on heuristics over fine-grained visual cues. While introducing reasoning capabilities significantly suppresses these errors, it introduces a paradox of over-thinking; \textit{Logical Impossibility} errors surge as models actively reject counter-intuitive visual truths, and \textit{Alignment Failures} rise, indicating a disconnect where instinctive language priors override the correct deductive chain during final generation.

\begin{figure}[h]
    \centering
    \includegraphics[width=\linewidth]{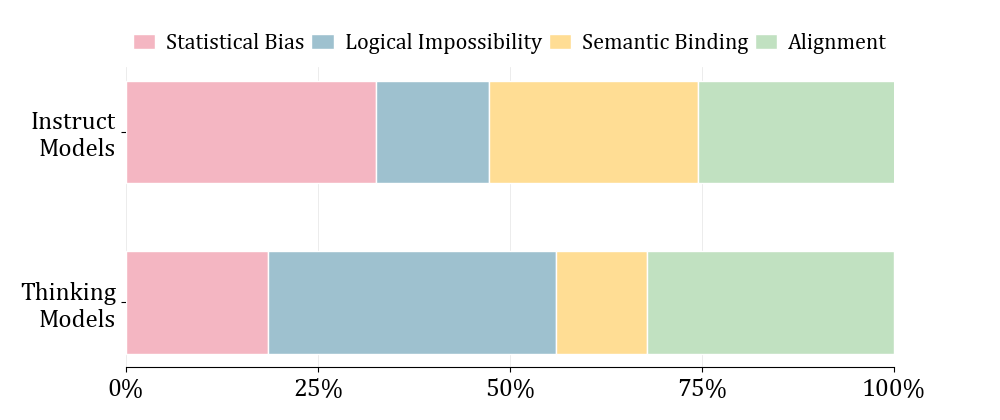} 
    \caption{Distribution of error types of open-source MLLMs. }
    \label{fig:error_dist}
\end{figure}

\subsection{CAIT Dataset Quality Evaluation}

To assess the quality of the CAIT dataset, we conducted two complementary evaluations: perceptual image quality using no-reference IQA metrics and semantic alignment using CLIP cosine similarity. This ensures model failures are due to reasoning deficits rather than image artifacts or misalignment.

\paragraph{Perceptual Quality (NR-IQA).} We computed NIQE and BRISQUE scores for CAIT images to quantitatively evaluate their perceptual naturalness and presence of artifacts. The CAIT dataset achieves a mean NIQE score of 3.08 and a mean BRISQUE score of 8.12. In comparison, images from the Winoground dataset exhibit lower quality, with a mean NIQE of 4.33 and a BRISQUE of 32.88. These results indicate that CAIT images possess superior perceptual quality and fewer structural distortions than Winoground, ensuring that model errors are not a byproduct of poor image signal quality.

\begin{table}[h]
\centering
\begin{tabular}{lcc}
\toprule
Dataset & NIQE $\downarrow$ & BRISQUE $\downarrow$ \\
\midrule
CAIT & 3.08 & 8.12 \\
Winoground & 4.33 & 32.88 \\
\bottomrule
\end{tabular}
\caption{Perceptual quality (NR-IQA) for CAIT and Winoground. Lower score indicate better image quality.}
\label{tab:iqa_scores}
\end{table}

\paragraph{Semantic Alignment (CLIP).} Using CLIP ViT-B/32, we computed cosine similarity between image embeddings and captions. CAIT images achieve a mean correct-caption score of 0.295 and reversed-caption score of 0.295, yielding a near-zero semantic margin of -0.0007. For Winoground, the corresponding scores are 0.261, 0.258, and a margin of 0.0031. These results confirm that CAIT images are semantically aligned, and the near-zero margin emphasizes that models cannot rely solely on object recognition but must reason about interactions and compositional logic.

\begin{table}[h]
\centering
\begin{tabular}{lcc}
\toprule
Dataset & Correct Caption & Reversed Caption \\
\midrule
CAIT & 0.295 & 0.295 \\
Winoground & 0.261 & 0.258 \\
\bottomrule
\end{tabular}
\caption{Mean CLIP cosine similarity between images and captions for CAIT and Winoground datasets.}
\label{tab:clip_alignment}
\end{table}
\section{Conclusion}



We introduce CAIT, a novel benchmark comprising 400 independent scenes designed to fill the current gap in counter-intuitive action reasoning. Our evaluation and analysis reveals a stark contrast: while human intuition and leading proprietary models perform exceptionally well, open-sourced models in System 1 mode perform at the chance level, indicating a strong reliance on language priors. 
Although System 2 reasoning provide notable performance gains, it slow down the responses and introduce new failure mode where models over-rationalize and explicitly reject visual truths.
Furthermore, we demonstrate that targeted fine-tuning and structured prompting can effectively mitigate this language bias, providing insights for developing future multimodal models capable of better physical and interaction logic understanding.


\section*{Limitations}
\paragraph{Monolingual Evaluation} Currently, our benchmark is restricted to English. It is important to verify that our findings may generalize across various languages and cultural contexts. Our future work will expand the benchmark to multilingual scenarios such as Chinese.
\paragraph{Restriction to Static Imagery} The scenes of our work are specifically within a single static frame. Extending to video-based counter-intuitive interactions would introduce dynamic challenges such as temporal and audio reasoning, providing a more comprehensive evaluation for MLLMs.

\section*{Acknowledgments}
The authors would like to thank the anonymous reviewers for their helpful suggestions and comments. This work was supported by the Brain Science and Brain-like Intelligence Technology --- National Science and Technology Major Project 2021ZD0204100 (2021ZD0204105 to N.D.).

\clearpage






\bibliography{custom}

\clearpage
\appendix
\section{Supplementary Materials}

\subsection{Examples of Verb Library Filtering}
\label{apx:verb_library_filtering}
To ensure the high quality and recognizability of the action scenes in CAIT, we followed four core principles during the filtering of the \textit{Sensitive-Action Verb Library}. Table \ref{tab:verb_examples} provides specific examples of verbs that passed or failed these criteria.

\begin{table*}[t]
\centering
\small
\begin{tabular}{lp{2.0cm}p{8.5cm}}
\toprule
\textbf{Criterion} & \textbf{Pass/Fail Case} & \textbf{Selection / Exclusion Logic} \\
\midrule
Semantic Consistency & salute/venerate & "Salute" is defined by specific, identifiable hand and eye movements. In contrast, abstract verbs like "venerate" lack a uniform visual manifestation, leading to low annotator agreement on the specific physical action. \\
\midrule
Semantic Non-interchangeability & chase/play with & "Play with" is semantically symmetrical; swapping the agent and patient does not fundamentally change the scene's logic. We prioritize asymmetric verbs like "chase," where role reversal creates a distinct counter-intuitive conflict. \\
\midrule
Static Recognizability & kick/think & The action must be discernible from a single static frame without temporal context. Cognitive states like "think" are internal and lack observable physical interaction markers, making them unsuitable for VQA tasks. \\
\midrule
Commonality & push/block & We select high-frequency, general-purpose verbs like "push" to ensure model evaluation focuses on logic rather than vocabulary. Specialized terms like "block" (e.g., in sports contexts) are excluded due to their heavy reliance on domain-specific knowledge. \\
\bottomrule
\end{tabular}
\caption{Detailed Selection and Exclusion Criteria for the Sensitive-Action Verb Library. This taxonomy ensures that all verbs used for counter-intuitive scene generation possess objective visual markers and clear logical asymmetry.}
\label{tab:verb_examples}
\end{table*}

\subsection{Detailed Information of Annotators}
\label{apx:annotator}

\paragraph{Annotator Selection}
The selected annotators represent diverse age groups and educational backgrounds, with a minimum requirement of an elementary school education and a fundamental understanding of real-world common knowledge. By excluding professional annotation training, we ensure that the baseline reflects natural human intuition rather than specialized task-solving strategies.

\paragraph{Annotation Process and Compensation}
\begin{itemize}
    \item \textbf{Interface:} Annotations were conducted via a dedicated web interface that presents questions as individual, sequential items.
    \item \textbf{Task Flow:} Annotators received one question at a time and submitted their responses directly on the webpage to ensure a focused, rapid judgment process.
    \item \textbf{Ethics and Compensation:} To ensure high-quality data and ethical labor standards, each annotator was compensated at a rate of 7 dollars per hour, approximately 2.5 times the local minimum wage.
\end{itemize}


\paragraph{Annotation Statistics} 
Table~\ref{apx:Human_annotation_statistics} shows detailed performance metrics for each annotator. The results demonstrate high accuracy and consistency across diverse groups, with an average Fleiss' $\kappa$ of 0.880. The average response time of 3.740 seconds per item aligns with the "blink-level" judgment criteria, confirming that the task successfully captures rapid human intuition.

\newcolumntype{Y}{>{\centering\arraybackslash}X}
\begin{table*}[t]
\centering
\small
\begin{tabularx}{\textwidth}{@{} l c X Y Y Y @{}} 
\toprule
\textbf{Annotator} & \textbf{Age} & \textbf{Description} & \textbf{Acc} & \textbf{F1} & \textbf{Avg Time (s)} \\
\midrule
A1 & 15 & Undergraduate student & 0.945 & 0.945 & 3.7 \\
A2 & 21 & Master student        & 0.942 & 0.942 & 3.3 \\
A3 & 22 & Master student        & 0.980 & 0.980 & 3.4 \\
A4 & 59 & General public        & 0.953 & 0.953 & 4.1 \\
A5 & 58 & General public        & 0.920 & 0.920 & 4.2 \\
\midrule
Mean & -- & -- & 0.948 & 0.948 & 3.740 \\
Std  & -- & -- & 0.020 & 0.020 & 0.404 \\
\midrule
\multicolumn{3}{l}{Inter-annotator agreement (Fleiss' $\kappa$)} & \multicolumn{3}{c}{0.88} \\
\bottomrule
\end{tabularx}
\caption{Human annotation statistics on the CAIT VQA task.}
\label{apx:Human_annotation_statistics}
\end{table*}

\subsection{Prompts of MLLMs during Inference}
\label{apx:prompts_for_MLLMs_during_Inference}
Table \ref{tab:original_prompt} shows the original prompt for MLLMs.
Table \ref{tab:format_instructions} shows the CoT prompt for Instruct Models
\begin{table*}[t]
\small
\centering
\enlargethispage{3\baselineskip} 

\begin{minipage}[t]{0.48\textwidth}
    \centering
    \begin{tabularx}{\linewidth}{@{}X@{}}
    \toprule
    Which of the following option better describes the image? \\
    Option A: \{option\_a\} \\
    Option B: \{option\_b\} \\
    \midrule
    \textbf{Important Note:} Answer directly with the option letter (A or B) only. \\
    \bottomrule
    \end{tabularx}
    \vspace{-6pt} 
    \captionof{table}{Original Prompt for MLLMs during inference.}
    \label{tab:original_prompt}

    \vspace{0.5em} 

    \begin{tabularx}{\linewidth}{@{}X@{}}
    \toprule
    Which of the following option better describes the image? \\
    Option A: \{option\_a\} \\
    Option B: \{option\_b\} \\
    \midrule
    Please begin with a brief analysis... \\
    \midrule
    \textbf{Important Note:} You must write... \\
    \bottomrule
    \end{tabularx}
    \vspace{-6pt}
    \captionof{table}{CoT instructions for Instruct Models.}
    \label{tab:format_instructions}
\end{minipage}
\hfill
\begin{minipage}[t]{0.48\textwidth}
    \centering
    \renewcommand{\arraystretch}{0.9} 
    \begin{tabularx}{\linewidth}{@{}X@{}}
    \toprule
    \textbf{1. Description:} Describe the content... \par
    The primary objects or scene in the image. \par
    Visual features such as colors, shapes, and textures. \par
    Any text, symbols, or logos. \par
    The overall layout and composition of the image. \\
    \midrule
    \textbf{2. Options:} Analyze the meaning... \par
    Option A: \{option\_a\} \par
    Option B: \{option\_b\} \\
    \midrule
    \textbf{3. Reasoning:} Compare the image content... \par
    Compare the image content with the description in Option B... \par
    Analyze which option describes the image content more accurately. \par
    Consider the possibility of uncommon... \\
    \midrule
    \textbf{4. Conclusion:} Based on your analysis... \par
    Summarize your reasoning process. \par
    Clearly state the reasons for choosing A or B. \par
    Provide the final answer. \\
    \midrule
    \textbf{Important Note:} At the very end... \\
    \bottomrule
    \end{tabularx}
    \vspace{-6pt}
    \captionof{table}{Detailed Structured Prompt for System 2 thinking mode.}
    \label{tab:structured_prompt}
\end{minipage}
\end{table*}

\subsection{Structured prompts for System 2 thinking mode}
Table \ref{tab:structured_prompt} shows the structured prompt for System 2 mode.
\label{apx:structured_prompt}

\subsection{Per-Judge Results of LLM-re-Eval}
\label{apx:llm_re_eval_judges}
The LLM-re-Eval column reported in Table \ref{tab:main_results} is the average over the three proprietary judges. Table \ref{tab:llm_re_eval} reports the accuracy obtained with each individual judge, showing that the ranking of the evaluated models is largely independent of the judge being used.

\begin{table*}[t]
\centering
\begin{tabular}{lccc}
\toprule
\textbf{Model} & \textbf{Gemini-2.5-pro} & \textbf{Claude-4.6-Opus} & \textbf{GPT-5.4} \\
\midrule
GT Prompt (upper bound) & 0.983 & 0.985 & 0.985 \\
\midrule
\multicolumn{4}{l}{\textit{Proprietary Models}} \\
Gemini-2.5-pro & 0.838 & 0.823 & 0.843 \\
Claude-4.6-Opus & 0.803 & 0.770 & 0.790 \\
\midrule
\multicolumn{4}{l}{\textit{Open-source Instruct MLLMs}} \\
Qwen3-VL-Instruct-8B & \textbf{0.773} & \textbf{0.758} & \textbf{0.755} \\
Qwen3-VL-Instruct-4B & 0.728 & 0.690 & 0.703 \\
Qwen2.5-VL-7B & 0.668 & 0.653 & 0.660 \\
LLaMA3.2-Vision-11B & 0.640 & 0.588 & 0.593 \\
Gemma-3-12B & 0.618 & 0.585 & 0.593 \\
Qwen2.5-VL-3B & 0.590 & 0.570 & 0.593 \\
LLaVA-NeXT-7B & 0.558 & 0.505 & 0.582 \\
Gemma-3-4B & 0.553 & 0.538 & 0.532 \\
\midrule
\multicolumn{4}{l}{\textit{Open-source Thinking MLLMs}} \\
GLM-4.1V-Thinking-9B & \textbf{0.738} & \textbf{0.718} & \textbf{0.743} \\
Qwen3-VL-Thinking-8B & 0.705 & 0.678 & 0.705 \\
Kimi-VL-A3B-Thinking-16B & 0.705 & 0.673 & 0.690 \\
Qwen3-VL-Thinking-4B & 0.693 & 0.683 & 0.710 \\
InternVL-3.5-4B & 0.688 & 0.663 & 0.698 \\
InternVL-3.5-8B & 0.683 & 0.620 & 0.663 \\
\bottomrule
\end{tabular}
\caption{Different Judges of LLM-re-Eval. Values denoted in \textbf{bold} indicate the highest performance in each open-source model group for each judge.}
\label{tab:llm_re_eval}
\end{table*}

\subsection{Performance of Models across different domains}
Table \ref{tab:acc_specific_model_across_domain} shows the performance of specific models across different domains.
\label{apx:acc_across_domains}

\subsection{Performance of different models using structured prompts}
\label{apx:acc_structured_prompts_on_different_models}
Table \ref{tab:instruct_structured_prompt} and Table \ref{tab:thinking_structured_prompt} show the performance of different models using structured prompts.

\begin{table*}[t]
\centering
\begin{tabular}{lccccc}
\toprule
\textbf{Model} & \textbf{Param} & \textbf{A-A} & \textbf{B-S} & \textbf{H-A} & \textbf{H-H} \\
\midrule
\multicolumn{6}{l}{\textbf{\textit{$\sim$10B Instruct Models}}} \\
Qwen2.5-VL-Instruct & 7B & 0.486 & 0.447 & 0.483 & 0.352 \\
Qwen3-VL-Instruct & 8B & 0.500 & 0.342 & 0.421 & 0.297 \\
Gemma-3-it & 12B & 0.389 & 0.263 & 0.434 & 0.290 \\
LLaMA3.2-Vision & 11B & 0.361 & 0.421 & 0.386 & 0.290 \\
LLaVA-NeXT & 7B & 0.500 & 0.474 & 0.441 & 0.386 \\
\midrule
\multicolumn{6}{l}{\textbf{\textit{$\sim$3B Instruct Models}}} \\
Qwen2.5-VL-Instruct & 3B & 0.528 & 0.368 & 0.428 & 0.338 \\
Qwen3-VL-Instruct & 4B & 0.500 & 0.263 & 0.317 & 0.269 \\
Gemma-3-it & 4B & 0.500 & 0.184 & 0.352 & 0.290 \\
\midrule
\multicolumn{6}{l}{\textbf{\textit{$\sim$10B Thinking Models}}} \\
Qwen3-VL-thinking & 8B & 0.653 & 0.579 & 0.641 & 0.648 \\
GLM-4.1V-Thinking & 9B & 0.653 & 0.579 & 0.600 & 0.641 \\
Kimi-VL-A3B-Thinking & 16B & 0.653 & 0.184 & 0.324 & 0.221 \\
Internvl-3.5 & 8B & 0.403 & 0.184 & 0.393 & 0.283 \\
\midrule
\multicolumn{6}{l}{\textbf{\textit{$\sim$3B Thinking Models}}} \\
Qwen3-VL-thinking & 4B & 0.611 & 0.684 & 0.600 & 0.641 \\
Internvl-3.5 & 4B & 0.569 & 0.553 & 0.517 & 0.621 \\
\bottomrule
\end{tabular}
\caption{Detailed performance of MLLMs across different domains (A-A: Animal-Animal, B-S: Bio-StillLife, H-A: Human-Animal, H-H: Human-Human).}
\label{tab:acc_specific_model_across_domain}
\end{table*}
\begin{table}[ht]
\centering
\small
\begin{tabular}{lccc}
\toprule
\textbf{Model} & \textbf{Param} & \textbf{Acc} & \textbf{F1 score} \\
\midrule
\multicolumn{4}{l}{\textbf{\textit{$\sim$10B Instruct MLLMs}}} \\
Qwen2.5-VL-Instruct & 7B & 0.538 & 0.534 \\
Qwen3-VL-Instruct & 8B & 0.750 & 0.750 \\
Gemma-3-it & 12B & 0.535 & 0.538 \\
LLama3.2-vision & 11B & 0.458 & 0.458 \\
LLaVA-NeXT & 7B & 0.490 & 0.450 \\
\midrule
\multicolumn{4}{l}{\textbf{\textit{$\sim$3B Instruct MLLMs}}} \\
Qwen2.5-VL-Instruct & 3B & 0.413 & 0.413 \\
Qwen3-VL-Instruct & 4B & 0.663 & 0.663 \\
Gemma-3-it & 4B & 0.473 & 0.440 \\
\bottomrule
\end{tabular}
\caption{Performance of Instruct MLLMs using structured prompt.}
\label{tab:instruct_structured_prompt}
\end{table}
\begin{table}[ht]
\centering
\small
\begin{tabular}{lccc}
\toprule
\textbf{Model} & \textbf{Param} & \textbf{Acc} & \textbf{F1 score} \\
\midrule
\multicolumn{4}{l}{\textbf{\textit{$\sim$10B Thinking MLLMs}}} \\
Qwen3-vl-thinking & 8B & 0.703 & 0.703 \\
GLM-4.1V-Thinking & 9B & 0.675 & 0.698 \\
Kimi-VL-A3B-Thinking & 16B & 0.733 & 0.733 \\
InternVL-3.5 & 8B & 0.610 & 0.610 \\
\midrule
\multicolumn{4}{l}{\textbf{\textit{$\sim$3B Thinking MLLMs}}} \\
Qwen3-vl-thinking & 4B & 0.720 & 0.720 \\
InternVL-3.5 & 4B & 0.580 & 0.575 \\
\bottomrule
\end{tabular}
\caption{Performance of Thinking MLLMs using structured prompt.}
\label{tab:thinking_structured_prompt}
\end{table}

\section{Prompt of Workflow Details}
\label{app:prompts}

We present the core instructions for the workflow in \cref{tab:prompt_step1,tab:prompt_step2,tab:prompt_step3,tab:prompt_step4}. This includes the task definition and the constraints provided to the models.

\begin{table}[t]
    \small
    \renewcommand{\arraystretch}{1.2}

    \begin{tabularx}{\linewidth}{>{\raggedright\arraybackslash}X}
        \toprule
        \textbf{\large Step 1: Semantic Scene Generation (Verb $\to$ Text)} \\
        \midrule
        \textbf{Model:} GPT-4o-mini \hfill \textbf{Temperature:} 0.7 \\
        \midrule
        \textbf{System Prompt:} \\
        Please note the following examples I have provided. Each group is divided into two parts: the normal scene and the scene that is logical but extremely uncommon after the master-guest inversion. When reversing, try to keep the general structure of the original sentence's language description unchanged, only swapping the positions of the subject and object. Following these examples, generate semantic scenarios that meet the requirements for the verbs I gave you, with 10 groups for each verb. \par
        \vspace{0.3em}
        \textit{(Few-shot examples omitted)} \par
        \vspace{0.4em}
        You MUST respond in valid json format. Output a json object with the following structure:
        \texttt{\{ "verb": "\textless input\_verb\textgreater", "scene\_sets": [ \{ "id": \textless number\textgreater, "common": ["\textless common\_sentence\textgreater"], "uncommon": ["\textless uncommon\_sentence\textgreater"] \} ] \}} \\
        \vspace{0.3em}
        \textbf{User Prompt:} \\
        \texttt{\{ "verb": "\textless verb\textgreater" \}} \\
        \midrule
        \textbf{Output:} \\
        \texttt{\{ "verb": "\textless verb\textgreater",} \\
        \texttt{\ \ "scene\_sets": [ \{ "id": 1, "common": ["\textless common\_sentence\textgreater"], "uncommon": ["\textless uncommon\_sentence\textgreater"] \}, ... ] \}} \\
        \bottomrule
    \end{tabularx}
    
    \caption{Prompt details for Step 1: Generating semantic scene pairs from verbs.}
    \label{tab:prompt_step1}
\end{table}

\begin{table}[t]
    \small
    \renewcommand{\arraystretch}{1.2}

    \begin{tabularx}{\linewidth}{>{\raggedright\arraybackslash}X}
        \toprule
        \textbf{\large Step 2: Human Verification } \\
        \midrule
        
        \textbf{Operator:} Human Expert \hfill \textbf{Interface:} Custom GUI\\
        \midrule

        \includegraphics[width=0.95\linewidth]{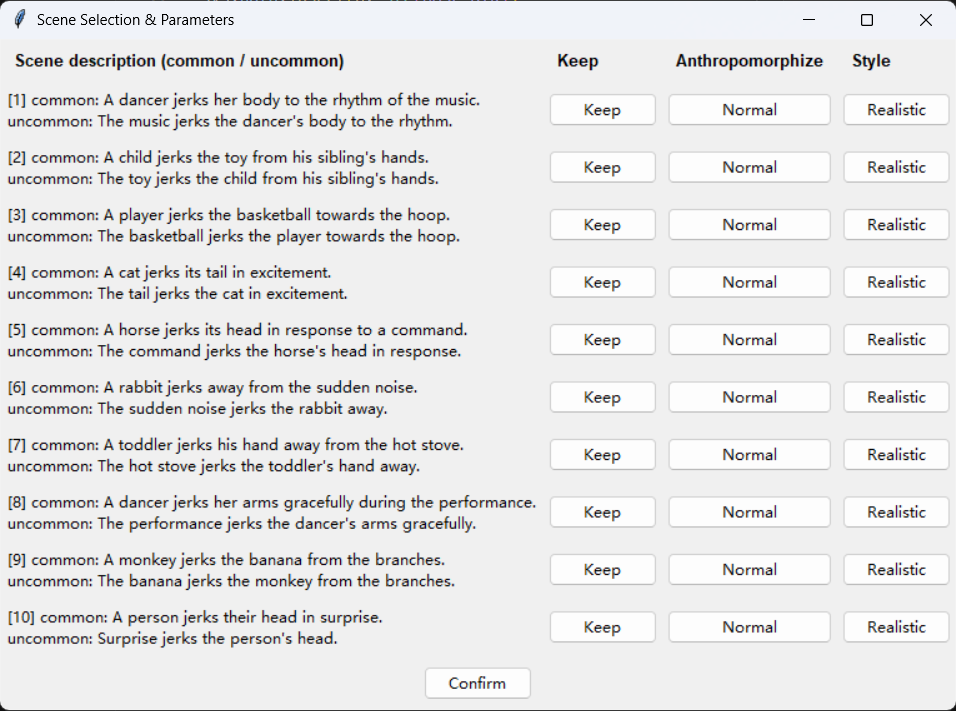} \\ 
        \textit{\footnotesize Figure A: The custom GUI for manual scene selection and parameter configuration.} \\
        
        \vspace{0.3em}
        \textbf{Process Description:} \\
        The raw semantic pairs generated in Step 1 are presented to a human annotator via a graphical user interface. The annotator performs quality control and parameter setting for each scene to ensure high usability for image generation. \par
        \vspace{0.3em}
        \textbf{Human Actions:} \\
        1. \textbf{Selection:} Review 10 generated scene pairs. Toggle the ``Keep'' button for scenes that maintain logical causality after inversion and filter out nonsensical ones. \\
        2. \textbf{Anthropomorphize:} Set the subject's nature (e.g., ``Normal'' vs. ``Anthropomorphic'') based on whether the inverted subject requires human-like traits to perform the action. \\
        3. \textbf{Style Setting:} Assign a visual style ( ``Realistic'' or ``Cartoon'') to guide the subsequent image generation. \\
        \midrule
        \textbf{Input Data:} \\
        \texttt{\{ "verb": "\textless verb\textgreater", "scene\_sets": [ ... (10 raw pairs) ... ] \}} \\
        \midrule
        \textbf{Output Data:} \\
        \texttt{\{ "verb": "\textless verb\textgreater",} \\
        \texttt{\ \ "scene\_sets": [ \{ "id": 1, "common": [...], "uncommon": [...], "selected": true, "anthropomorphize": true/false, "style": "realistic/cartoon" \}, ... ] \}} \\
        \bottomrule
    \end{tabularx}
    
    \caption{Details for Step 2: Human-in-the-loop selection and parameter configuration.}
    \label{tab:prompt_step2}
\end{table}

\begin{table}[t]
    \small
    \renewcommand{\arraystretch}{1.2}
    \begin{tabularx}{\linewidth}{>{\raggedright\arraybackslash}X}
        \toprule
        \textbf{\large Step 3: Scene Description Generation (Text $\to$ Description)} \\
        \midrule
        \textbf{Model:} GPT-4o-mini \hfill \textbf{Temperature:} 0.7 \\
        \midrule
        \textbf{System Prompt:} \\
        You are given:
        \texttt{- verb: \{verb\}, - scene\_sets: \{scene\_sets\_json\}} \par
        \vspace{0.3em}
        Task: For each uncommon semantic scenario (subject--object inversion) in the scene\_sets, generate ONE highly detailed and visually grounded description suitable for Gemini's text-to-image model. \par
        Each description must:
        1. Clearly convey the subject--object role reversal described in the uncommon\_text.
        2. Respect the per-scene options (anthropomorphize, style).
        3. Be vivid and concrete, describing what the viewer can see. \par
        \vspace{0.3em}
        Output strictly in json format. You MUST return a valid json object with the following structure:
        \texttt{\{ "verb": "\textless verb\textgreater", "records": [ \{ "scene\_id": 1, "uncommon\_image\_description": "\textless description\textgreater", ... \} ] \}} \par
        \vspace{0.3em}
        \textbf{User Prompt:} \\
        \texttt{\{ "verb": "\textless verb\textgreater", "scene\_sets": [ \{ "id": 1, "uncommon": ["\textless uncommon\_sentence\textgreater"], ... \}, ... ] \}} \\
        \midrule
        \textbf{Output:} \\
        \texttt{\{ "verb": "\textless verb\textgreater", "records": [ \{ "scene\_id": 1, "uncommon\_text": "\textless uncommon\_sentence\textgreater", "uncommon\_image\_description": "\textless uncommon\_image\_description\textgreater", "style": "realistic/cartoon" \}, ... ] \}} \\
        \bottomrule
    \end{tabularx}

    \caption{Prompt details for Step 3: Converting text pairs into detailed visual descriptions.}
    \label{tab:prompt_step3}
\end{table}

\begin{table}[t]
    \small
    \renewcommand{\arraystretch}{1.2}
    \begin{tabularx}{\linewidth}{>{\raggedright\arraybackslash}X}
        \toprule
        \textbf{\large Step 4: Paradoxical Image Generation (Description $\to$ Image)} \\
        \midrule
        \textbf{Model:} Gemini-2.5-flash-image \hfill \textbf{Temperature:} 0.7 \\
        \midrule
        \textbf{Full Prompt Template:} \\
        You are a visual reasoning model that deliberately violates common sense.
        You must prioritize paradox, contradiction, and reversed causality over realism. \par
        \vspace{0.3em}
        World rules:
        1. Physical laws can be inverted.
        2. Causes may appear after results.
        3. Object scale, weight, and function may contradict logic.
        4. The image must look realistic in style but impossible in logic.
        5. Do not explain the paradox visually -- let it appear naturally. \par
        \vspace{0.4em}
        Here is the uncommon and counter-intuitive scene you have to generate: \par
        \texttt{\textless uncommon\_image\_description\textgreater} \par
        \vspace{0.3em}
        Image size: 512$\times$512 pixels \\
        \midrule
        \textbf{Output:} \\
        \textit{[Generated Image: A 512$\times$512 pixel paradoxical image visualizing the description]} \\
        \bottomrule
    \end{tabularx}

    \caption{Prompt details for Step 4: Generating paradoxical images from descriptions.}
    \label{tab:prompt_step4}
\end{table}

\section{Error Analysis}
\label{sec:error_analysis}

To better understand the limitations of current MLLMs, we conduct a qualitative error analysis. We random sample 200 error cases from the Structured Prompt experiments of both Instruct models and Thinking models. We categorize the typical mistakes into several types, ranging from perceptual errors to logical inconsistencies in action sequences.

\subsection{Case Studies}
We list several representative cases below. For each case, we provide the image, the model's response, and a brief analysis of why the model failed to correctly interpret the counter-intuitive action.
Specific error cases are presented in \cref{tab:case1_error_analysis,tab:case2_error_analysis,tab:case3_error_analysis,tab:case4_error_analysis}. Specifically, \cref{tab:case1_error_analysis} illustrates the \textit{Statistical Frequency Bias} error. \Cref{tab:case2_error_analysis} demonstrates \textit{Logical/Physical Impossibility}. \Cref{tab:case3_error_analysis} shows \textit{Spatial/Semantic Binding Failure}, and \cref{tab:case4_error_analysis} presents an example of \textit{Alignment/Extraction Failure}.
\begin{table}[h]
    \centering
    \small 
    \renewcommand{\arraystretch}{1.3} 
    \begin{tabularx}{\linewidth}{>{\raggedright\arraybackslash}X}
        \toprule
        \textbf{CASE 1: Prior Bias Overriding Visual Evidence (Thinking Model - InternVL-3.5-8B)} \\
        \midrule
        \textbf{QUESTION:} \\
        \begin{minipage}[t]{0.6\linewidth}
            \vspace{0pt} 
            Choose the description that best matches the image: \\
            A. a little boy is guiding a parent \\
            B. a parent is guiding a little boy 
        \end{minipage}
        \hfill 
        \begin{minipage}[t]{0.35\linewidth}
            \vspace{0pt}
            \includegraphics[width=0.95\linewidth]{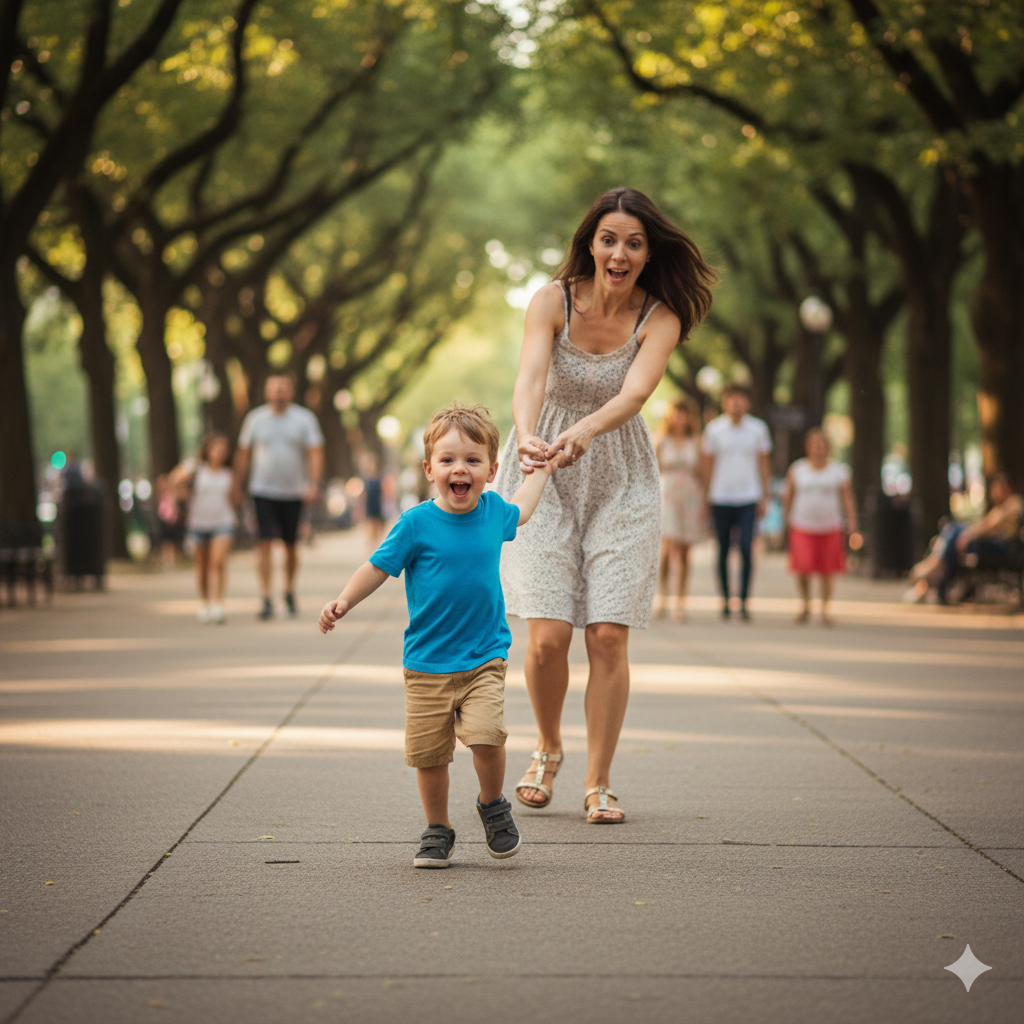} 
        \end{minipage}
        \\ 
        \midrule
        \textbf{MODEL THINKING:} \\
        Okay, let's tackle this step by step. First, I need to analyze the image as per the instructions. The image shows a woman and a young boy holding hands and running down a tree-lined path. The woman is smiling broadly, and the boy looks happy too. [...] \\
        The key here is to determine who is leading whom. The woman is holding the boy's hand, and they're both moving forward. The boy is slightly ahead, but the woman is supporting him by holding his hand. \textcolor{red}{\textbf{Typically, when a parent guides a child, they might hold the child's hand and lead them, especially if the child is running.}} The boy's position isn't necessarily leading the woman; instead, they seem to be moving together with the woman supporting him. So, the parent (woman) is guiding the child (boy) as they both move forward. That would make option B more accurate. However, sometimes the child might take the lead, but in this case, the woman's posture and the way they're holding hands suggest she's guiding him. \textcolor{red}{\textbf{The boy is smiling and seems to be enjoying the run, but the guiding action is more about the parent leading the child.}} So, I think B is the correct answer here. \\
        \textbf{Final Answer: B} \\
        \midrule
        \textbf{FAILURE ANALYSIS:} \\
                \textbf{Acknowledgement of Evidence:} The model initially acknowledges the valid visual cue, noting that ``The boy is slightly ahead.'' \\
                
                \textbf{Activation of Priors:} However, it immediately counters this observation with a probabilistic generalization derived from its training data: ``Typically, when a parent guides a child... they lead them.'' \\
                
                \textbf{Rejection of Reality:} Driven by this statistical likelihood, the model rejects the low-probability visual reality (a child guiding a parent). It rationalizes the scene to fit the high-probability distribution, arguing the boy is ``not necessarily leading,'' effectively hallucinating a scenario that aligns with its internal priors rather than the image. \\
        \bottomrule
    \end{tabularx}
    \caption{Error Analysis Case 1: Type 1 -- Statistical Frequency Bias}
    \label{tab:case1_error_analysis}
\end{table} 
\begin{table}[h]
    \centering
    \small 
    \renewcommand{\arraystretch}{1.3} 
    \begin{tabularx}{\linewidth}{>{\raggedright\arraybackslash}X}
        \toprule
        \textbf{CASE 2: Logical Impossibility (Instruction Model - Gemma3-12B-it)} \\
        \midrule
        \textbf{QUESTION:} \\
        \begin{minipage}[t]{0.6\linewidth}
            \vspace{0pt} 
            Choose the description that best matches the image: \\
            A. A cat swats a toy mouse. \\
            B. A toy mouse swats the cat. 
        \end{minipage}%
        \hfill
        \begin{minipage}[t]{0.35\linewidth}
            \vspace{0pt}
            \includegraphics[width=0.95\linewidth]{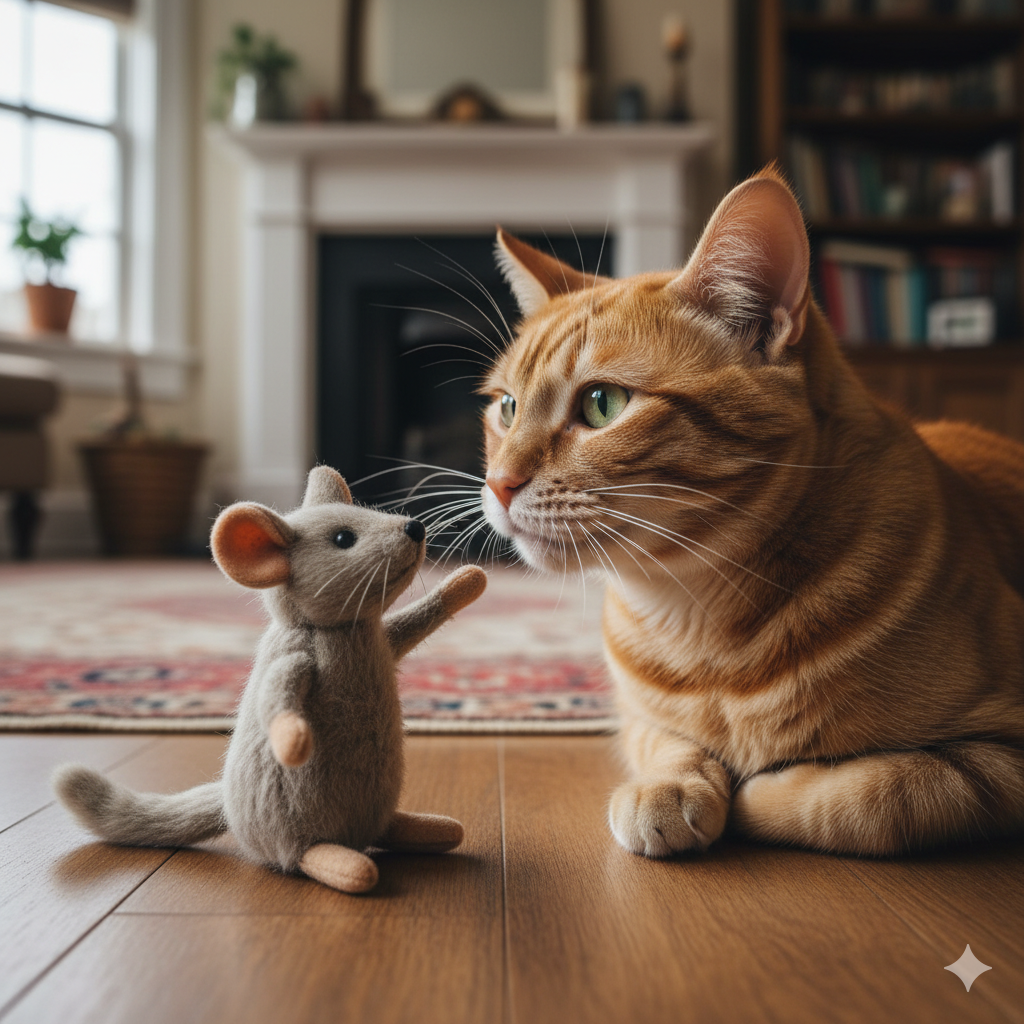} 
        \end{minipage}
        \\ 
        \midrule
        \textbf{REASONING TRACE:} \\
        Okay, let's analyze the image and the provided options... The image depicts a domestic cat and a toy mouse... \\
        Let's compare the image content with each option... Option A: The image clearly shows the cat with its paw extended... The cat is the active agent... \\
        Option B: The image does \textit{not} depict the toy mouse striking the cat. \textcolor{red}{\textbf{The toy mouse is stationary and passive. The concept of a toy mouse ``swatting'' a cat is physically impossible and defies the typical understanding of the relationship between a cat and its toys.}} It's an unconventional and highly implausible scenario... \\
        My reasoning process involved a detailed description of the image, followed by a comparison of the image content with each option. Option A, "A cat swats a toy mouse," accurately reflects the scene, as the cat is actively engaging with the toy. \textcolor{red}{\textbf{Option B is simply not possible... Therefore, Option A is the correct choice.}} \\
        \textbf{Final Answer: A} \\
        \midrule
        \textbf{FAILURE ANALYSIS:} \\
        
        \textbf{Acknowledgement of Entities:} The model correctly identifies the scene's components (``domestic cat,'' ``toy mouse'') and their spatial proximity. \\
        
        \textbf{Activation of Constraints:} However, it invokes a strict logical rule regarding inanimate objects, explicitly labeling the scenario as ``physically impossible'' and ``highly implausible'' because toys cannot initiate action. \\
        
        \textbf{Rejection of Reality:} Instead of accepting the staged or humorous nature of the image (where a toy might visually ``swat'' a cat), the model rigidly rejects the visual evidence. It forces the interpretation to align with real-world physics, concluding that the event is ``simply not possible'' despite the visual cue. \\
        \bottomrule
    \end{tabularx}
    \caption{Error Analysis Case 2: Type 2 -- Logical Impossibility}
    \label{tab:case2_error_analysis}
\end{table} 
\begin{table}[h]
    \centering
    \small 
    \renewcommand{\arraystretch}{1.3} 
    \begin{tabularx}{\linewidth}{>{\raggedright\arraybackslash}X}
        \toprule
        \textbf{CASE 3: Semantic Binding Failure (Instruct Model - Gemma3-12B-it)} \\
        \midrule
        \textbf{QUESTION:} \\
        \begin{minipage}[t]{0.6\linewidth}
            \vspace{0pt} 
            Choose the description that best matches the image: \\
            A. A knight stabs a dragon with a sword. \\
            B. A dragon stabs the knight with a sword. \\
        \end{minipage}%
        \hfill
        \begin{minipage}[t]{0.35\linewidth}
            \vspace{0pt}
            \includegraphics[width=0.95\linewidth]{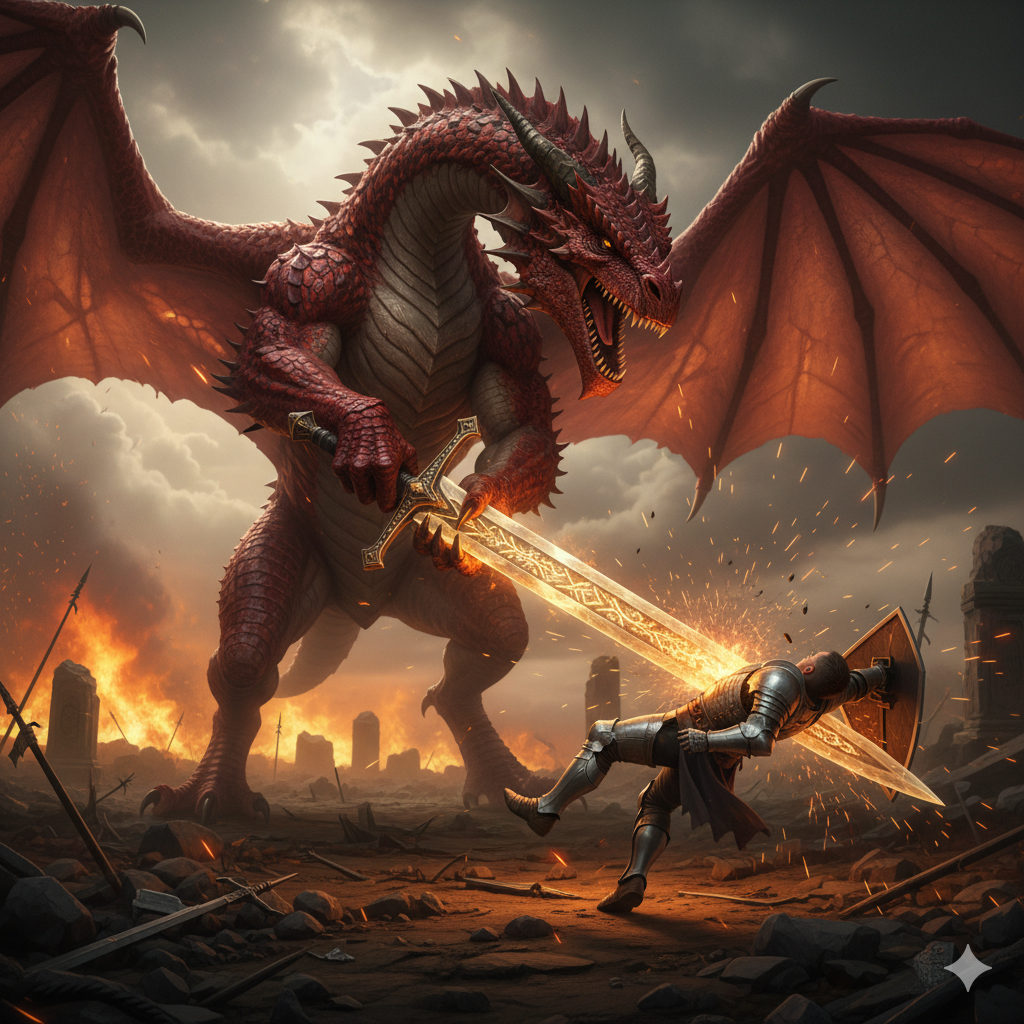} 
        \end{minipage}
        \\ 
        \midrule
        \textbf{REASONING TRACE:} \\
        Okay, let's analyze the image... The image depicts a dramatic, fantastical scene of a knight battling a large, red dragon... \\
        Let's compare the image content with each option... Option A (Knight stabs dragon): The knight \textit{is} holding a sword raised, and it appears to be aimed at the dragon. The sword is close to the dragon's body. The angle of the sword and the knight's posture suggest an attack. This aligns well with the image. \\
        Option B (Dragon stabs knight): \textcolor{red}{\textbf{The dragon is holding a sword, but the sword is pointed downwards, away from the knight... there's no indication that the dragon is actively stabbing him.}} The dragon's posture is more of a defensive or aggressive stance, rather than a stabbing motion. This does not align well with the image. \\
        While it's possible to interpret the scene in various ways, the most straightforward interpretation is that the knight is attempting to strike the dragon. \textcolor{red}{\textbf{Therefore, the image more accurately represents a knight attempting to stab a dragon.}} \\
        \textbf{Final Answer: A} \\
        \midrule
        \textbf{FAILURE ANALYSIS:} \\
        
        \textbf{Object Identification:} The model correctly detects all key elements in the scene: the knight, the dragon, and the swords. \\
        
        \textbf{Attribute Misbinding:} However, it fails to correctly bind the action to the agent. It misinterprets the spatial orientation of the dragon's sword (acknowledging it is ``pointed downwards'' but failing to recognize the attack) while incorrectly binding the offensive action to the knight's posture, effectively ignoring the sword in the dragon's hand. \\
        
        \textbf{Subject-Object Reversal:} Consequently, the model swaps the semantic roles of the entities. It assigns the ``agent'' role (stabbing) to the knight and the ``patient'' role to the dragon, directly contradicting the visual evidence where the dragon is the aggressor. \\
        \bottomrule
    \end{tabularx}
    \caption{Error Analysis Case 3: Type 3 -- Semantic Binding Failure}
    \label{tab:case3_error_analysis}
\end{table} 
\begin{table}[h]
    \centering
    \small 
    \renewcommand{\arraystretch}{1.3} 
    \begin{tabularx}{\linewidth}{>{\raggedright\arraybackslash}X}
        \toprule
        \textbf{CASE 4: Alignment Failure (Thinking Model - InternVL-3.5-8B)} \\
        \midrule
        \textbf{QUESTION:} \\
        \begin{minipage}[t]{0.6\linewidth}
            \vspace{0pt} 
            Choose the description that best matches the image: \\
            A. a father is guiding a girl riding a bike on the road \\
            B. a girl on the road is guiding her father riding a bike 
        \end{minipage}%
        \hfill 
        \begin{minipage}[t]{0.35\linewidth}
            \vspace{0pt}
            \includegraphics[width=0.95\linewidth]{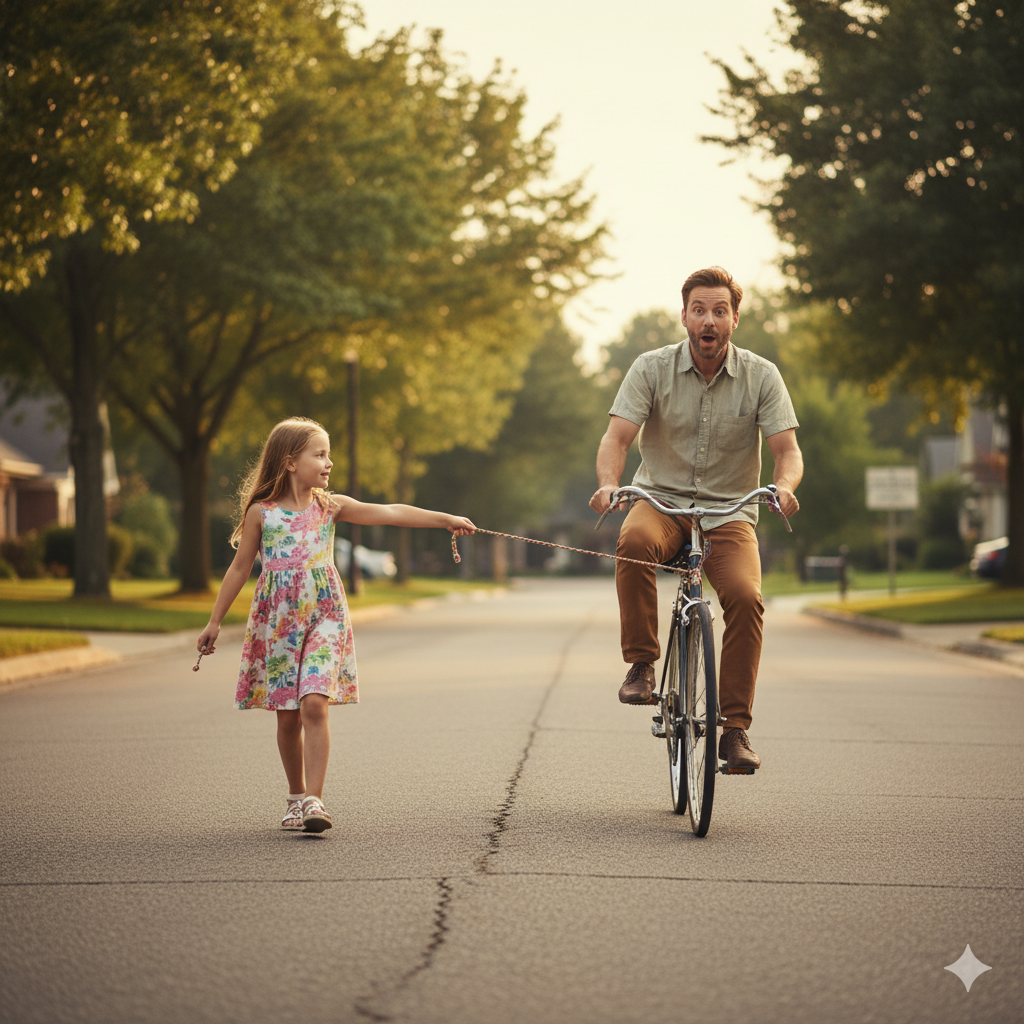} 
        \end{minipage}
        \\ 
        \midrule
        \textbf{MODEL THINKING:} \\
        Okay, let's tackle this step by step. [...] The scene shows a man riding a bicycle on a residential street, and a young girl walking beside him. The girl is holding a rope attached to the bike's handlebars. [...] Option A says a father is guiding a girl riding a bike. \textbf{But in the image, the girl isn't on the bike; she's walking alongside.} So Option A might be incorrect. [...] The key is who is riding the bike. In the image, the father is on the bike. Therefore, Option A is incorrect because the girl isn't on the bike. Option B is the only remaining choice, even if it's unconventional. [...] \textbf{Therefore, the correct answer must be Option B, where the girl is guiding her father who is riding the bike.} Even though it's unconventional, the image shows the girl holding the rope attached to the bike's handlebars, which could imply she's steering it, thus guiding the father. \\
        \textcolor{red}{\textbf{Final Answer: A}} \\
        \midrule
        \textbf{FAILURE ANALYSIS:} \\
                \textbf{Correct Visual Reasoning:} Unlike other types of errors, the model correctly perceives the visual reality in its chain-of-thought, explicitly stating ``the girl isn't on the bike'' and identifying that Option A is factually incorrect. \\
                
                \textbf{Final Answer Generation Collapse:} Despite deducing the correct answer (B) internally, the model fails at the final generation step. The explicit reasoning is discarded in favor of a hallucinated label. \\
                
                \textbf{Re-emergence of Priors:} This disconnect highlights a specific alignment failure. The strong textual prior associated with ``father guiding girl'' (common) versus ``girl guiding father'' (rare) overwhelms the logical conclusion during the token decoding phase, causing the model to revert to the statistical mean despite its own correct reasoning. \\
        \bottomrule
    \end{tabularx}
    \caption{Error Analysis Case 4: Type 4 -- Alignment Failure}
    \label{tab:case4_error_analysis}
\end{table}


\end{document}